\documentclass[manuscript,screen]{acmart}

\usepackage{graphicx}
\usepackage{amsmath}
\usepackage{algpseudocode}
\usepackage{booktabs}
\usepackage{multirow}
\usepackage{subcaption}

\def\ie{\textit{i.e.}}
\def\eg{\textit{e.g.}}
\def\cf{\textit{c.f.}}
\definecolor{mygray}{gray}{.9}
\definecolor{lightpink}{RGB}{255, 222, 235}
\definecolor{cellred}{RGB}{249,211,244}
\definecolor{cellgreen}{RGB}{211,249,216}
\definecolor{mygreen}{RGB}{93,173,85}
\definecolor{rred}{RGB}{190,0,0}
\definecolor{lightblue}{RGB}{208,235,255}

\newcommand{\pub}[1]{{\color{gray}{\tiny{[{#1}]}}}}

\newcommand{\tabstyle}[4]{ 
    \centering
    \resizebox{#1\textwidth}{!}{
    \setlength\tabcolsep{#2pt}
    \renewcommand\arraystretch{#3}
    #4
    }
}

\AtBeginDocument{%
  }

\setcopyright{acmlicensed}
\copyrightyear{2018}
\acmYear{2018}
\acmDOI{XXXXXXX.XXXXXXX}
\acmConference[Conference acronym 'XX]{Make sure to enter the correct
  conference title from your rights confirmation email}{June 03--05,
  2018}{Woodstock, NY}
\acmISBN{978-1-4503-XXXX-X/2018/06}

\begin{document}

\title{Velocity-coupled Representation Refinement for Satellite Orbit Prediction}

\author{Yue Yang}
\email{yueyang05@xjtu.edu.cn}
\affiliation{%
  \institution{Xi'an Jiaotong University}
  \city{Xi'an}
  \country{China}
}

\author{Zhiqiang Wu}
\email{3125358293@stu.xjtu.edu.cn}
\affiliation{%
  \institution{Xi'an Jiaotong University}
  \city{Xi'an}
  \country{China}
}

\author{Saiyu Qi}
\email{saiyu-qi@mail.xjtu.edu.cn}
\affiliation{%
  \institution{Xi'an Jiaotong University}
  \city{Xi'an}
  \country{China}
}

\author{Fan Ma}
\correspondingauthor
\email{mafan@zju.edu.cn}
\affiliation{%
  \institution{Zhejiang University}
  \city{Hangzhou}
  \country{China}
}

\renewcommand{\shortauthors}{Yang et al.}


\begin{abstract}
Satellite orbit prediction, which aims to forecast future orbital trajectories from historical observations, is of great importance to collision warning and safe space operations.
With recent advances in time-series forecasting, learning-based methods have emerged as a promising end-to-end solution for satellite prediction.
In orbital dynamics, a satellite state is typically described by both position and velocity, where position characterizes the spatial geometry of the trajectory and velocity reflects its instantaneous direction and rate of change.
However, most existing methods mainly focus on temporal dependencies within position sequences, \ie, learning future trajectory evolution from historical orbital positions, while rarely exploiting the intrinsic coupling between position and velocity, which is essential for modeling satellite motion.
To this end, we propose \textsc{OrbitNet}, a velocity-aware representation learning method for accurate satellite orbit prediction.
It lifts conventional position-sequence forecasting to a position-velocity coupled representation learning paradigm, by explicitly exploiting the relationships among satellite state variables.
Specifically, we develop a velocity-coupled representation refinement strategy to enhance positional representations through cross-variable interactions between position and velocity.
In addition, we introduce orbital segment modeling, which partitions historical trajectories into temporal segments and performs segment-level temporal learning, thereby capturing local motion variations and long-range evolution patterns.
Extensive experiments show that \textsc{OrbitNet} outperforms large time-series foundation models (\eg, TimesFM and Times-MoE), as well as representative general forecasting methods (\eg, DLinear, iTransformer, and DropPatch), under both in-domain evaluation on Starlink and zero-shot evaluation across six unseen satellite constellations.
We expect this work to encourage further exploration of satellite-aware representation learning for trajectory time-series forecasting.
\end{abstract}

\ccsdesc[500]{Mathematics of computing~Time series analysis}
\ccsdesc[500]{Computing methodologies~Neural networks}


\maketitle

\section{Introduction}
\label{sec:introduction}

Satellite orbit prediction, aiming to forecast future orbital trajectories from historical observations, is a fundamental yet challenging task in satellite monitoring and space situational awareness.
As large satellite constellations continue to expand and the near-Earth orbital environment becomes increasingly congested, accurate orbit prediction has become essential for downstream orbital operations, including collision avoidance~\cite{reiland2021assessing,caldas2024precise} and space traffic management~\cite{Zhang2024SelfsimilarTP,Fan2025SatelliteEI}.

Conventional orbit prediction methods mainly follow a model-driven paradigm, where future orbital states are inferred from historical observations using predefined orbit propagators.
SGP4~\cite{hoots1980models} is a representative tool that utilizes two-line element (TLE) records to estimate future orbital states.
Although such physics-based approaches are interpretable, their predictive capability is often constrained by handcrafted dynamical assumptions and unmodeled perturbations.
With the rapid development of deep learning, time-series forecasting has witnessed substantial progress in both prediction accuracy and computational efficiency, ranging from classical recurrent architectures~\cite{sahoo2019large,guo2023multivariate,lin2025segrnn} to modern Transformer-based paradigms~\cite{wu2021autoformer,zhou2021informer,itransformer,fedformer}.
Such progress opens up a data-driven route to satellite orbit prediction, where nonlinear orbital evolution patterns can be learned directly from large-scale historical observations.

Along this direction, existing learning-based orbit prediction methods can be roughly categorized into two lines.
The first line mainly focuses on architectural design by, for instance, utilizing recurrent networks~\cite{Chu2025KiGRUFL,shin2022selective,fu2025multi} or incorporating attention mechanisms~\cite{jeong2024decomposed} to better capture temporal dependencies in orbital sequences.
The other line is tailored to TLE-oriented prediction~\cite{caldas2024precise,Zhang2025ASM,marie2025tle}, where models are trained to predict future TLE records, which are then used by orbit propagators for subsequent orbital state computation.

\begin{figure}[t]
    \centering
    \includegraphics[width=0.70\linewidth]{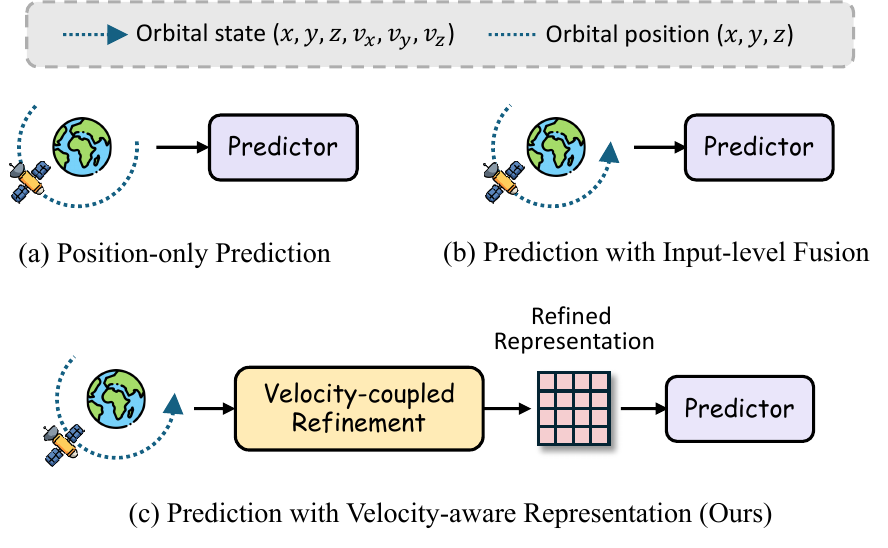}
    \caption{Satellite orbit prediction paradigms: (a) Position-only prediction uses historical orbital positions $(x,y,z)$ as input.
    (b) Input-level fusion appends velocity variables $(v_x,v_y,v_z)$ to the position sequence.
    (c) \textsc{OrbitNet} performs velocity-coupled refinement to enhance positional representations with velocity-aware motion cues.}
    \Description{Comparison of three satellite orbit prediction paradigms: position-only input, direct position-velocity input fusion, and OrbitNet's velocity-coupled refinement of positional representations.}
    \label{fig:intro}
\end{figure}

Despite the progress of existing orbit prediction methods, most approaches still rely on position-only inputs, which may limit their ability to fully characterize orbital evolution (Figure~\ref{fig:intro}(a)), leading to suboptimal predictive performance.
From the perspective of orbital mechanics, position and velocity are not two independent variables, but two coupled components of the orbital state.
To be specific, position describes the spatial geometry of the trajectory, while velocity characterizes its instantaneous direction and rate of change.
Their interaction jointly determines how the orbit evolves over time.
Hence, a pivotal question naturally raises: \textit{what should an effective orbital representation encode for accurate orbit prediction?}
Ideally, it should not only i) preserve the positional information of the historical trajectory, but also ii) be well structured to encode velocity-aware motion cues.
One straightforward solution is to concatenate velocity with position as the full orbital-state input, as shown in Figure~\ref{fig:intro}(b).
However, such input-level fusion treats velocity as an auxiliary variable and leaves position-velocity interactions to be learned implicitly, which may underexploit the motion cues carried by velocity states.
These observations motivate us to move beyond position-only prediction and input-level fusion, and instead rethink orbit prediction from a representation learning perspective.

To this end, we propose \textsc{OrbitNet}, a forecasting framework designed for accurate orbit prediction. 
The core idea of \textsc{OrbitNet} is to enhance positional sequence learning by explicitly injecting velocity-aware motion cues into the positional representation space, rather than merely appending velocity as an additional input channel.
Specifically, we introduce a velocity-coupled refinement strategy, which captures cross-variable interactions between position and velocity to refine positional representations in a velocity-aware manner.
In this way, the resulting representation remains aligned with the position space, preserving the spatial structure of historical positions while incorporating motion cues from velocity states for future trajectory forecasting.
In addition, we develop orbital segment modeling to partition the trajectory sequence into temporal segments for patch-wise temporal learning.
By learning representations at the segment level, \textsc{OrbitNet} captures local motion variations within individual segments while preserving long-range dependencies across the historical trajectory.
Extensive experiments demonstrate that \textsc{OrbitNet} achieves superior performance over existing large time-series foundation models (\eg, TimesFM~\cite{das2023decoder} and Times-MoE~\cite{shi2024time}), as well as leading general forecasting approaches (\eg, DLinear~\cite{dlinear}, iTransformer~\cite{itransformer}, and DropPatch~\cite{droppatch}).
Such advantages are consistently observed in both in-domain forecasting and challenging zero-shot evaluation across six unseen satellite constellations, validating the effectiveness and generalization capability of the proposed representation learning framework.

In a nutshell, our contributions are three-fold:

\begin{itemize}

    \item We revisit satellite orbit prediction from a representation learning perspective and highlight the importance of enhancing positional representations with velocity-aware motion cues for accurate orbit prediction.

    \item We propose \textsc{OrbitNet}, a forecasting framework that integrates velocity-coupled refinement with orbital segment modeling. It enhances positional representations through velocity-aware cross-variable interactions and captures orbital evolution patterns via patch-wise temporal learning.

    \item We conduct extensive experiments under both in-domain and zero-shot settings across multiple satellite constellations. The results demonstrate that \textsc{OrbitNet} consistently outperforms large time-series foundation models and leading general forecasting approaches, while exhibiting strong cross-constellation generalization.

\end{itemize}

\section{Related Work}
\label{sec:related}

In this section, we briefly review recent advances in satellite orbit prediction (Section~\ref{subsec:orbit_prediction}) and then discuss representative deep learning methods for time-series forecasting (Section~\ref{subsec:time_series_forecasting}).

\subsection{Satellite Orbit Prediction}
\label{subsec:orbit_prediction}

Satellite orbit prediction is a long-standing problem in aerospace engineering, with the goal of forecasting future orbital positions from historical observations.
It plays an essential role in a wide range of downstream applications, including collision avoidance~\cite{Yu2024SafeRL,Wang2025DataModelHS,Sun2022SatelliteFC} and space traffic management~\cite{Zhang2024SelfsimilarTP,Fan2025SatelliteEI}.
Traditional solutions mainly rely on physics-based propagators, such as the SGP4 model~\cite{hoots1980models}, which propagate orbital states by integrating equations of motion under simplified dynamical assumptions.
Although these model-driven approaches are computationally efficient and interpretable, their prediction accuracy often degrades due to uncertain initial conditions and unmodeled perturbations, \eg, atmospheric drag and solar radiation pressure.

To alleviate these limitations, recent studies have explored data-driven and hybrid paradigms for orbit prediction.
A common strategy is to combine physical propagation with machine learning-based correction, where learning models estimate the discrepancy between physics-based predictions and ground-truth trajectories~\cite{caldas2024machine,kazemi2024orbit,caldas2024precise,liu2024dual}.
Early attempts along this direction employ support vector machines~\cite{peng2018exploring} or Gaussian processes~\cite{peng2021fusion} to learn propagation errors and improve prediction accuracy.
Beyond correcting propagation errors, neural sequence models have also been introduced to directly learn orbital evolution from historical observations~\cite{xiao2026hybrid,xu2024digital,jeong2024decomposed,ren2019research,yonglong2025fdlstm,shin2022selective}.
These methods mainly focus on designing recurrent architectures~\cite{Chu2025KiGRUFL,shin2022selective,fu2025multi}, attention mechanisms~\cite{jeong2024decomposed}, or decomposed temporal modules~\cite{Lee2025SpatioTA} to better capture temporal dependencies in orbital sequences.
For example, \cite{jeong2024decomposed} devises a decomposed attention segment RNN to enhance dependency learning over orbital sequences, while \cite{shin2022selective} develops a selective tensorized multi-layer LSTM architecture for compact and effective recurrent sequence modeling.
In parallel, several studies focus on TLE-oriented prediction~\cite{caldas2024precise,Zhang2025ASM,marie2025tle}, where future TLE records or TLE-related errors are predicted and then used for subsequent orbital state computation.

Despite these advances, existing learning-based orbit prediction methods mainly focus on either improving temporal architectures or adapting neural networks to TLE-based prediction settings.
The representation structure of orbital state variables themselves remains less explored, especially the intrinsic coupling between position and velocity.
In contrast, our work studies satellite orbit prediction from a representation learning perspective and explicitly refines positional representations with velocity-aware motion cues for accurate future trajectory forecasting.

\subsection{Deep Learning for Time-series Forecasting}
\label{subsec:time_series_forecasting}

Over the past decade, deep learning has substantially advanced time-series forecasting, with backbone architectures evolving from recurrent models to Transformer-based frameworks.
In particular, recurrent architectures, \eg, RNN~\cite{elman1990finding} and LSTM~\cite{hochreiter1997long}, established the foundation for modern forecasting models by showing that temporal dependencies can be learned in an end-to-end manner.
Building on this paradigm, subsequent studies further improved forecasting performance by extracting multi-resolution temporal contexts~\cite{lian2008multiscale,yan2024multi,bi2026patchfusionmlp}, incorporating non-local dependency modeling~\cite{zeng2021topological,guo2023multivariate,fang2023stwave,hu2025pattern,li2024sa2e}, and introducing trend decomposition strategies~\cite{stitsyuk2025xpatch,dlinear,li2021modeling}.
These designs enrich temporal representations from different perspectives and have become important components of modern forecasting models.

Accordingly, Transformer-based solutions~\cite{zhou2021informer,wu2021autoformer,fedformer,itransformer,lai2024lightcts,tipirneni2022self} have attracted increasing attention in this field.
Benefiting from self-attention, these fully attentive paradigms can model long-range dependencies and achieve strong performance across various benchmarks.
Specifically, Informer~\cite{zhou2021informer} devises ProbSparse self-attention to reduce the quadratic time and memory complexity of standard attention.
Autoformer~\cite{wu2021autoformer} leverages series decomposition and auto-correlation for long-term forecasting, while FEDformer~\cite{fedformer} captures informative temporal representations in the frequency domain.
In parallel, MLP-based and linear forecasting models~\cite{bi2026patchfusionmlp,chen2023tsmixer,ekambaram2023tsmixer,yi2023frequency,zhang2026mdmlp,dlinear} emerge as lightweight yet competitive alternatives.
Instead of relying on recurrence or self-attention, these methods capture temporal dependencies through simple and effective mixing or decomposition operations, achieving favorable trade-offs between predictive accuracy and computational efficiency.

Recently, time-series foundation models have further broadened the scope of forecasting by learning general-purpose temporal representations from large-scale time-series corpora.
Representative models, such as TTM~\cite{ekambaram2024tiny}, TimesFM~\cite{das2023decoder}, MOIRAI~\cite{woo2024unified}, and Times-MoE~\cite{shi2024time}, show promising transfer ability across diverse forecasting scenarios.
Nevertheless, most existing forecasting models are designed for generic temporal sequences (\eg, weather~\cite{wu2021autoformer} and electricity~\cite{zhou2021informer}) and usually treat variables as ordinary input channels.
They therefore do not explicitly encode the position-velocity coupling that is central to satellite orbit prediction.
In contrast, our work develops a domain-aware forecasting framework that refines positional representations with velocity-aware motion cues for accurate orbit prediction.

\section{Methodology}
\label{sec:methodology}

In this section, we first formulate the satellite orbit prediction problem and clarify the representation principle behind our design.
We then present the proposed \textsc{OrbitNet} framework in Section~\ref{subsec:framework}, followed by its detailed network architecture in Section~\ref{subsec:architecture}.

\begin{figure*}[t]
    \centering
    \includegraphics[width=0.98\linewidth]{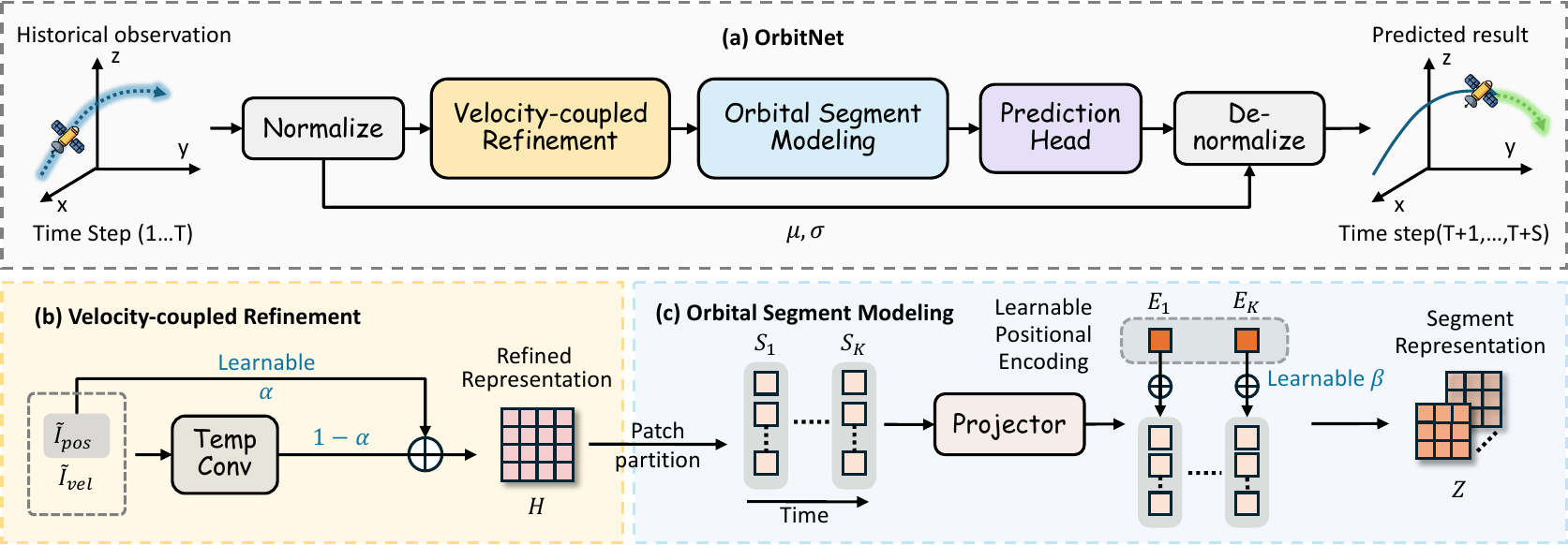}
    \caption{Overview of \textsc{OrbitNet}. Given historical position and velocity sequences, \textsc{OrbitNet} first refines positional representations through velocity-coupled refinement, where velocity-aware cross-variable correlations are injected into the positional representation space. It then partitions the refined representation into temporal segments for patch-wise temporal learning and predicts future orbital positions after denormalization.}    
    \Description{Architecture of OrbitNet, showing position and velocity sequences entering velocity-coupled representation refinement, followed by temporal segmentation, patch-wise temporal learning, and future position prediction.}
    \label{fig:framework}
\end{figure*}

\subsection{Method Overview}
\label{subsec:overview}

\subsubsection{Problem Setup}
\label{subsubsec:problem_setup}

Let a historical orbital observation be represented by a sequence of orbital states
$\mathcal{I}=(\mathbf{I}_{\mathrm{pos}}, \mathbf{I}_{\mathrm{vel}})$,
where
$\mathbf{I}_{\mathrm{pos}}=[\mathbf{p}_1,\mathbf{p}_2,\ldots,\mathbf{p}_T]\in\mathbb{R}^{P\times T}$
is the historical position sequence, and
$\mathbf{I}_{\mathrm{vel}}=[\mathbf{v}_1,\mathbf{v}_2,\ldots,\mathbf{v}_T]\in\mathbb{R}^{V\times T}$
denotes the corresponding velocity sequence.
Here, $T$ is the lookback window length, and $P$ and $V$ represent the dimensions of the position and velocity states, respectively.
In this work, we set $P\!=\!3$ and $V\!=\!3$, corresponding to the spatial coordinates $(x,y,z)$ and the instantaneous velocities $(v_x,v_y,v_z)$.

The goal of satellite orbit prediction is to forecast the future position sequence over the subsequent $S$ time steps, \ie,
$\widehat{\mathbf{Y}}=[\widehat{\mathbf{p}}_{T+1}, \widehat{\mathbf{p}}_{T+2}, \ldots, \widehat{\mathbf{p}}_{T+S}]\in\mathbb{R}^{P\times S}$,
with the ground truth denoted as $\mathbf{Y}\in\mathbb{R}^{P\times S}$.
The task is therefore formulated as forecasting future positions conditioned on both historical position and velocity sequences:

\begin{equation}
    \widehat{\mathbf{Y}}=\mathcal{F}(\mathbf{I}_{\mathrm{pos}}, \mathbf{I}_{\mathrm{vel}};\theta),
\end{equation}
where $\mathcal{F}(\cdot)$ is the orbit predictor and $\theta$ represents the learnable parameters.

\subsubsection{Our Idea}
\label{subsubsec:motivation}

The key question behind our design is how to construct an effective orbital representation for future position prediction.
As discussed in Section~\ref{sec:introduction}, such a representation should preserve the positional information of historical trajectories while encoding velocity-aware motion cues.
With this insight, \textsc{OrbitNet} treats position as the primary prediction-relevant signal and uses velocity to refine positional representations, rather than simply concatenating velocity with position as additional input channels.
Specifically, velocity information is injected into the positional representation space through cross-variable interaction, producing a velocity-enhanced positional representation for future trajectory forecasting.

\subsection{\textsc{OrbitNet}}
\label{subsec:framework}

As illustrated in Figure~\ref{fig:framework}, \textsc{OrbitNet} is built upon two core designs, \ie, velocity-coupled refinement and orbital segment modeling.
Specifically, \textsc{OrbitNet} first constructs a velocity-aware positional representation from the joint position-velocity input, then models orbital evolution through patch-wise temporal learning, and finally forecasts future orbital positions with a lightweight prediction head.

\subsubsection{Velocity-coupled Refinement}
\label{subsubsec:augmentation}

A straightforward way to incorporate velocity information is to directly concatenate velocity vectors with orbital positions as model inputs.
However, such input-level fusion treats position and velocity as separate channels and leaves their interactions to be learned implicitly by the predictor.
As a result, the motion cues carried by velocity states may not be fully exploited.
To address this limitation, we introduce velocity-coupled refinement, which injects velocity-aware correlation patterns into the positional representation space.

Formally, after instance-wise normalization, let
$\widetilde{\mathbf{I}}_{\mathrm{pos}}\in\mathbb{R}^{P\times T}$
and
$\widetilde{\mathbf{I}}_{\mathrm{vel}}\in\mathbb{R}^{V\times T}$
denote the normalized position and velocity sequences, respectively.
We first concatenate them along the channel dimension and apply a one-dimensional temporal convolution to extract a correlation-aware representation $\mathbf{H}_{\mathrm{corr}}$ via:

\begin{equation}
    \mathbf{H}_{\mathrm{corr}} =    f_{\mathrm{conv}}\big([\widetilde{\mathbf{I}}_{\mathrm{pos}};\widetilde{\mathbf{I}}_{\mathrm{vel}}]\big),
\end{equation}
where $[\cdot;\cdot]$ is the channel-wise concatenation, resulting in a multivariate sequence in $\mathbb{R}^{(P+V)\times T}$.
The operator $f_{\mathrm{conv}}(\cdot)$ captures temporal correlation patterns across position and velocity channels.
Its output $\mathbf{H}_{\mathrm{corr}}\in\mathbb{R}^{P\times T}$ remains aligned with the positional representation space, enabling velocity-aware cues to refine position-based forecasting.

To retain the fundamental positional signal while incorporating velocity-aware motion cues, we adaptively fuse $\mathbf{H}_{\mathrm{corr}}$ with the normalized position sequence by:

\begin{equation}
    \mathbf{H} = \alpha \widetilde{\mathbf{I}}_{\mathrm{pos}} + (1-\alpha)\mathbf{H}_{\mathrm{corr}},
\label{eq:couple}
\end{equation}

where $\alpha\in[0,1]$ is a learnable scalar parameter that controls the contributions of the positional signal and the velocity-coupled representation.
Consequently, the refined representation $\mathbf{H}\in\mathbb{R}^{P\times T}$ preserves the spatial structure of historical positions and is refined by velocity-aware correlation patterns.
This representation is subsequently fed into orbital segment modeling for temporal representation learning.

\subsubsection{Orbital Segment Modeling}
\label{subsubsec:segment_modeling}

After deriving the refined representation $\mathbf{H}$, we further characterize its temporal evolution through orbital segment modeling.
Direct point-wise modeling over the full sequence may overlook orbital changes within local time windows.
We therefore partition the sequence into non-overlapping temporal segments and perform patch-wise temporal learning.
This design allows the model to capture local orbital variations within individual segments while maintaining a global view of the historical trajectory.

Specifically, for $\mathbf{H}\in\mathbb{R}^{P\times T}$, we first apply zero-padding along the temporal dimension and then divide the padded sequence into $K$ non-overlapping temporal segments:

\begin{equation}
    \{\mathbf{S}_1, \mathbf{S}_2, \ldots, \mathbf{S}_K\} = \texttt{Patchify}(\mathbf{H}),
\label{eq:patch}
\end{equation}
where each segment $\mathbf{S}_k\in\mathbb{R}^{P\times L_p}$ corresponds to a local temporal patch of length $L_p$, and $K=\lceil T/L_p \rceil$ is the total number of segments.
Each segment is then projected into a latent space by:

\begin{equation}
    \mathbf{z}_k=f_{\mathrm{proj}}(\mathbf{S}_k),
\label{eq:project}
\end{equation}
where $f_{\mathrm{proj}}(\cdot)$ denotes the segment projection operator, and $\mathbf{z}_k\in\mathbb{R}^{P\times d}$ is the representation of the $k$-th temporal segment.
This projection maps each local patch into a compact latent representation that summarizes its local orbital evolution pattern.
By stacking these representations along the segment dimension, we obtain
$\mathbf{Z}_{\mathrm{seg}}=[\mathbf{z}_1,\mathbf{z}_2,\ldots,\mathbf{z}_K]$,
where $\mathbf{Z}_{\mathrm{seg}}\in\mathbb{R}^{P\times K\times d}$.

To preserve temporal order across segments, we introduce a learnable segment positional embedding $\mathbf{E}_{\mathrm{seg}}\in\mathbb{R}^{P\times K\times d}$
and combine it with $\mathbf{Z}_{\mathrm{seg}}$:

\begin{equation}
    \mathbf{Z} = \beta \mathbf{Z}_{\mathrm{seg}} + (1-\beta)\mathbf{E}_{\mathrm{seg}},
\end{equation}
where $\beta\in[0,1]$ is a learnable scalar that adaptively regulates the influence of segment semantics and temporal positional priors.
The integrated representation $\mathbf{Z}$ captures both intra-segment motion trends and inter-segment temporal dependencies, and is then passed to the prediction head for future position prediction.

\subsection{Detailed Network Architecture}
\label{subsec:architecture}

Our \textsc{OrbitNet} consists of five major components (\cf~Figure~\ref{fig:framework}):

\begin{itemize} 

    \item \textit{Normalization and Denormalization}, which normalizes each input orbital sequence before representation learning and restores the predicted positions to the original physical scale after forecasting.
    Specifically, we adopt an instance-wise normalization scheme~\cite{kim2021reversible} to normalize the input sequence along the temporal dimension using the sample-specific mean ${\mu}$ and standard deviation ${\sigma}$.
    The final prediction is denormalized with the corresponding positional statistics.

    \item \textit{Velocity-coupled Refinement}, which constructs the refined representation $\mathbf{H}$ by refining positional representations with velocity-aware cross-variable correlations.

    \item \textit{Orbital Segment Modeling}, which partitions the refined representation $\mathbf{H}$ into non-overlapping temporal patches and maps them into a latent space through the segment projection operator $f_{\mathrm{proj}}$.
    A learnable segment positional embedding is further incorporated to preserve temporal order across patches.

    \item \textit{Prediction Head}, transforming the learned temporal representation into the target forecasting horizon and generates the predicted future position sequence $\widehat{\mathbf{Y}}$.

    \item \textit{Training Loss}, $\mathcal{L}_{\mathrm{pred}}$, which optimizes the entire network in an end-to-end manner.
    In practice, the training objective is defined as
    $\mathcal{L}_{\mathrm{pred}}=\left\|\widehat{\mathbf{Y}}-\mathbf{Y}\right\|_2^2$,
    where $\mathbf{Y}$ denotes the ground-truth future position sequence.

\end{itemize}

\section{Experiments}
\label{sec:experiments}

In this section, we evaluate the effectiveness of the proposed \textsc{OrbitNet}. We first describe the dataset curation process and the experimental protocol in Sections~\ref{subsec:dataset_curation} and~\ref{subsec:experimental_setup}, respectively.
We then report the in-domain forecasting results on Starlink in Section~\ref{subsec:main_results}, followed by zero-shot evaluation across six unseen satellite constellations in Section~\ref{subsec:zero_shot_evaluation}.
Section~\ref{subsec:qualitative_results} provides qualitative trajectory visualization results.
In Section~\ref{subsec:ablation_study}, we conduct ablation studies to validate the effectiveness of the key designs in \textsc{OrbitNet}.
We further evaluate the long-term forecasting capability of \textsc{OrbitNet} in Section~\ref{subsec:long_term_prediction}. Finally, Section~\ref{subsec:correlation_analysis} presents a variable correlation analysis to interpret the learned cross-variable dependencies.

\subsection{Dataset Curation}
\label{subsec:dataset_curation}

\begin{table*}[t]
\centering
\caption{Summary of the satellite orbit datasets (Section~\ref{subsec:dataset_curation}).}

\label{tab:dataset_summary}
\tabstyle{0.92}{8}{1.2}{

\begin{tabular}{ccccc}
\toprule
Dataset & Satellite Type & Setting & \# Satellites & Time Span \\

\midrule
Starlink  & Communication      & \texttt{In-domain Train/Test} & 260 & 2019.11.14--2025.02.25 \\

ASTROCAST & IoT                & \texttt{Zero-shot Test}       & 15  & 2025.02.26--2026.03.13 \\

CAPELLA   & Remote Sensing     & \texttt{Zero-shot Test}       & 8   & 2025.02.26--2026.03.13 \\

ICEYE     & Remote Sensing     & \texttt{Zero-shot Test}       & 52  & 2025.02.26--2026.03.13 \\

LEMUR     & Earth Observation  & \texttt{Zero-shot Test}       & 69  & 2025.02.26--2026.03.13 \\

SKYSAT    & Earth Observation  & \texttt{Zero-shot Test}       & 15  & 2025.02.26--2026.03.13 \\

KINEIS    & Communication      & \texttt{Zero-shot Test}       & 25  & 2025.02.26--2026.03.13 \\
\bottomrule
\end{tabular}
}
\end{table*}

We construct the satellite orbit forecasting datasets through the following procedure. Detailed dataset statistics are summarized in Table~\ref{tab:dataset_summary}.

\begin{itemize}
  
    \item \textit{Raw TLE collection.}
    We construct satellite orbit forecasting datasets from publicly accessible two-line element (TLE) records collected from the Space-Track platform~\cite{spacetrack2025documentation}. These records provide orbital element information for satellites across different constellations and serve as the source data for subsequent state sequence generation.

    \item \textit{State sequence generation.}
    Raw TLE updates are typically sparse and irregularly sampled, making them unsuitable for direct use in fixed-window time-series forecasting. To obtain uniformly sampled orbital trajectories, we use the Orekit library~\cite{orekit2026} to convert discrete TLE records into orbital state sequences sampled at a $1$-minute interval. Each state contains both position and velocity variables, \ie, $(x,y,z,v_x,v_y,v_z)$, which are used as the input variables in this study.

    \item \textit{Dataset split.}
    The dataset statistics are summarized in Table~\ref{tab:dataset_summary}. We use the Starlink constellation as the source domain for model training. In total, $260$ Starlink satellites are collected, among which $208$ and $52$ satellites are used for in-domain training and testing, respectively. The corresponding observation period spans from $2019$-$11$-$14$ to $2025$-$02$-$25$. To evaluate cross-constellation generalization, we further collect data from six unseen constellations, including ASTROCAST, CAPELLA, ICEYE, KINEIS, LEMUR, and SKYSAT. These zero-shot datasets cover the period from $2025$-$02$-$26$ to $2026$-$03$-$13$ and are used only for testing.

\end{itemize}

\subsection{Experimental Setup}
\label{subsec:experimental_setup}

\paragraph{Training}

Following common practice in time-series forecasting~\cite{dlinear,fedformer,itransformer}, we transform each uniformly sampled orbital trajectory into non-overlapping forecasting samples.
Specifically, the historical look-back window is set to $T{=}512$, the prediction horizon is set to $S{=}90$, and the sliding stride is fixed to $T+S$.
Thus, each input sample contains $512$ historical time steps with six orbital state variables, while the predicted target is the future $90$-step position sequence over $(x,y,z)$.
For optimization, we adopt Adam~\cite{adam} with an initial learning rate of $0.001$ and use CosineAnnealingLR~\cite{loshchilov2016sgdr} as the learning rate scheduler to promote stable convergence during training.

\paragraph{Testing}

We evaluate \textsc{OrbitNet} under both \texttt{in-domain} and \texttt{zero-shot} settings.
For \texttt{in-domain} evaluation, the model is trained and tested on the Starlink dataset using the predefined satellite-level $8{:}2$ split.
For \texttt{zero-shot} evaluation, the model trained on Starlink is directly applied to six unseen satellite constellations without any fine-tuning.
These two protocols allow us to assess both prediction accuracy within the source constellation and generalization ability across different satellite constellations.

\paragraph{Baselines}

We compare \textsc{OrbitNet} with several widely recognized forecasting models, including three Transformer-based methods (AutoFormer~\cite{wu2021autoformer}, FEDformer~\cite{fedformer}, and iTransformer~\cite{itransformer}), two linear-based methods (DLinear~\cite{dlinear} and TimeXer~\cite{wang2024timexer}), two MLP-based methods (DropPatch~\cite{droppatch} and WPMixer~\cite{murad2025wpmixer}), and one CNN-based method (TimesNet~\cite{timesnet}).
Beyond these conventional baselines, we further include representative time-series foundation models, \ie, TTM~\cite{ekambaram2024tiny}, TimesFM~\cite{das2023decoder}, MOIRAI~\cite{woo2024unified}, and Times-MoE~\cite{shi2024time}, to provide a broader comparison.
To compare with domain-specific orbit prediction methods, we also include two orbit-specific learning baselines, \ie, KiGRU~\cite{Chu2025KiGRUFL} and DASR~\cite{jeong2024decomposed}.
For fair comparison, all learning-based baselines are evaluated under the same sequence forecasting protocol, \ie, they take the same six-dimensional historical state sequence $(x,y,z,v_x,v_y,v_z)$ as input and predict the same future three-dimensional position sequence $(x,y,z)$.
We note that \textsc{OrbitNet} is categorized as a linear-based forecasting model.

\paragraph{Evaluation Metrics}

As conventions, root mean squared error (RMSE) and mean absolute error (MAE) are employed to evaluate forecasting accuracy.
All errors are computed after denormalization and reported in meters for physical interpretability.
In particular, the reported results are averaged over the three spatial coordinates.

\subsection{Main Results}
\label{subsec:main_results}

\begin{table}
\centering
\caption{Performance comparison on the Starlink dataset (Section~\ref{subsec:main_results}). The best and second-best results are highlighted in \textcolor{red}{\textbf{red}} and \textcolor{blue}{\underline{blue}}, respectively.}

\label{tab:main_results}
\tabstyle{0.62}{8}{1.2}{
\begin{tabular}{rlccc}
\toprule
\multicolumn{2}{c}{Method} & Params & MAE & RMSE \\
\midrule

TTM~\cite{ekambaram2024tiny}\!\!&\!\!\pub{NeurIPS24} & 16,344,891  & 692.81 & 715.93 \\
TimesFM~\cite{das2023decoder}\!\!&\!\!\pub{ICML24} & 498,828,960 & 34.31 & 55.95 \\
MOIRAI~\cite{woo2024unified}\!\!&\!\!\pub{ICML24} & 15,855,786  & 95.78 & 174.13 \\
Times-MoE~\cite{shi2024time}\!\!&\!\!\pub{ICLR25} & 113,352,192 & 1297.16 & 1521.36 \\
\midrule

AutoFormer~\cite{wu2021autoformer}\!\!&\!\!\pub{NeurIPS21} & 10,505,217 & 63.49 & 107.51 \\
TimesNet~\cite{timesnet}\!\!&\!\!\pub{ICLR23} & 1,254,432 & 15.48 & \textcolor{blue}{\underline{22.92}} \\
FEDformer~\cite{fedformer}\!\!&\!\!\pub{ICML22} &290,194 & 162.72 & 186.09 \\
DLinear~\cite{dlinear}\!\!&\!\!\pub{AAAI23} & 554,040 & 8.27 & 46.15 \\
iTransformer~\cite{itransformer}\!\!&\!\!\pub{ICLR24} & 2,360,192 & 21.01 & 60.00 \\
TimeXer~\cite{wang2024timexer}\!\!&\!\!\pub{NeurIPS24} & 4,421,456 & 41.82 & 51.77 \\
DropPatch~\cite{droppatch}\!\!&\!\!\pub{AAAI25} & 425,100 & \textcolor{blue}{\underline{6.50}} & 50.99 \\
WPMixer~\cite{murad2025wpmixer}\!\!&\!\!\pub{AAAI25} & 6,964,577 & 163.77 & 185.45 \\
\midrule

DASR~\cite{jeong2024decomposed}\!\!&\!\!\pub{KDD24} & 433,718 & 153.48 & 235.85 \\
KiGRU~\cite{Chu2025KiGRUFL}\!\!&\!\!\pub{SMC25} & 1,582,023 & 96.63 & 160.79 \\
\midrule

\multicolumn{2}{c}{\textsc{OrbitNet}~(Ours)} & 78,432 & \textcolor{red}{\textbf{3.34}} & \textcolor{red}{\textbf{22.80}} \\
\bottomrule
\end{tabular}
}
\end{table}

Table~\ref{tab:main_results} reports the quantitative comparison on the Starlink dataset. As can be seen, the proposed \textsc{OrbitNet} achieves the best performance among all compared methods, including time-series foundation models, Transformer-based baselines, and domain-specific orbit prediction methods.
In particular, compared with the second-best MAE achieved by DropPatch~\cite{droppatch} ($6.50$ m), \textsc{OrbitNet} reduces the prediction error by nearly half, \ie, $48.6\%$.
In terms of RMSE, \textsc{OrbitNet} also achieves the best result, slightly reducing the error from $22.92$ m to $22.80$ m compared with the second-best baseline, TimesNet~\cite{timesnet}.
Compared with Transformer-based methods, \textsc{OrbitNet} exhibits a clear performance advantage, surpassing AutoFormer~\cite{wu2021autoformer} by $94.7\%$ and FEDformer~\cite{fedformer} by $97.9\%$ in MAE.
Compared with domain-specific orbit prediction methods, \textsc{OrbitNet} also achieves clear gains, reducing the MAE from $153.48$ m for DASR~\cite{jeong2024decomposed} and $96.63$ m for KiGRU~\cite{Chu2025KiGRUFL} to $3.34$ m, and lowering the RMSE from $235.85$ m and $160.79$ m to $22.80$ m.
Notably, even when compared with large-scale foundation models (\eg, Times-MoE~\cite{shi2024time}), which possess substantially larger model capacity, \textsc{OrbitNet} still achieves a significant performance margin while maintaining a smaller model size.

Beyond predictive precision, \textsc{OrbitNet} delivers superior accuracy with only $78,432$ parameters, standing as a compact model among all competitors.
For instance, compared with lightweight baselines, \eg, DropPatch ($425,100$ parameters) and DLinear ($554,040$ parameters), \textsc{OrbitNet} is approximately $5.4\times$ and $7.0\times$ smaller, respectively.
It is also more compact than the domain-specific baselines, with about $5.5\times$ fewer parameters than DASR and $20.2\times$ fewer parameters than KiGRU.
These results indicate that the proposed method not only improves prediction accuracy but also provides a parameter-efficient architecture, which is desirable for resource-constrained satellite onboard environments.

\subsection{Zero-shot Evaluation}
\label{subsec:zero_shot_evaluation}

\begin{table*}
\centering

\caption{Zero-shot performance on unseen satellite constellations (Section~\ref{subsec:zero_shot_evaluation}). The best results are highlighted in \textcolor{red}{\textbf{red}}, and the second-best results are underlined in \textcolor{blue}{blue}.}

\label{tab:zero_shot_results}
\tabstyle{1.0}{4}{1.2}{
\begin{tabular}{rlcccccccccccc}
\toprule
\multicolumn{2}{c}{\multirow{2}{*}{Method}} & \multicolumn{2}{c}{ASTROCAST} & \multicolumn{2}{c}{ICEYE} & \multicolumn{2}{c}{CAPELLA} & \multicolumn{2}{c}{KINEIS} & \multicolumn{2}{c}{LEMUR} & \multicolumn{2}{c}{SKYSAT} \\
\cmidrule(lr){3-4}\cmidrule(lr){5-6}\cmidrule(lr){7-8}\cmidrule(lr){9-10}\cmidrule(lr){11-12}\cmidrule(lr){13-14}
& & MAE & RMSE & MAE & RMSE & MAE & RMSE & MAE & RMSE & MAE & RMSE & MAE & RMSE \\
\midrule

TTM~\cite{ekambaram2024tiny}\!\!&\!\!\pub{NeurIPS24}
& 344.76 & 460.98
& 487.69 & 649.12
& 614.92 & 796.85
& 756.25 & 965.29
& 507.31 & 677.75
& 434.55 & 568.11 \\

TimesFM~\cite{das2023decoder}\!\!&\!\!\pub{ICML24}
& 27.82 & 47.64
& 35.76 & 50.89
& 62.59 & 82.95
& 125.48 & 155.79
& 34.16 & \textcolor{blue}{\underline{54.31}}
& 31.45 & 49.40 \\

MOIRAI~\cite{woo2024unified}\!\!&\!\!\pub{ICML24}
& 69.74 & 107.42
& 103.72 & 160.50
& 147.52 & 206.76
& 200.79 & 280.16
& 107.91 & 162.94
& 87.46 & 132.14 \\

Times-MoE~\cite{shi2024time}\!\!&\!\!\pub{ICLR25}
& 1279.35 & 1487.94
& 1193.91 & 1495.72
& 1265.79 & 1515.69
& 1244.32 & 1528.71
& 1184.93 & 1503.23
& 1194.36 & 1496.39 \\
\midrule

AutoFormer~\cite{wu2021autoformer}\!\!&\!\!\pub{NeurIPS21}
& 155.60 & 188.18
& 161.40 & 185.88
& 156.51 & 177.99
& 145.96 & 175.47
& 154.76 & 185.53
& 159.49 & 190.81 \\

TimesNet~\cite{timesnet}\!\!&\!\!\pub{ICLR23}
& 333.73 & 409.32
& 339.42 & 411.55
& 309.24 & 360.75
& 1450.89 & 1923.61
& 336.28 & 413.35
& 340.22 & 444.15 \\

FEDformer~\cite{fedformer}\!\!&\!\!\pub{ICML22}
& 48.10 & 401.92
& 46.02 & 400.95
& 160.76 & 386.21
& 48.75 & 379.57
& 79.81 & 425.24
& 46.81 & 401.11 \\

DLinear~\cite{dlinear}\!\!&\!\!\pub{AAAI23}
& \textcolor{blue}{\underline{12.35}} & \textcolor{red}{\textbf{15.81}}
& 19.19 & 48.58
& 29.84 & 55.77
& 787.42 & 1018.15
& 28.94 & 151.10
& \textcolor{blue}{\underline{11.15}} & \textcolor{blue}{\underline{15.17}} \\

TimeXer~\cite{wang2024timexer}\!\!&\!\!\pub{NeurIPS24}
& 17.47 & 25.50
& 46.49 & 64.58
& 91.98 & 116.06
& 154.63 & 183.66
& 43.35 & 68.99
& 20.25 & 44.38 \\

iTransformer~\cite{itransformer}\!\!&\!\!\pub{ICLR24}
& 12.71 & 51.38
& \textcolor{blue}{\underline{16.41}} & 63.01
& \textcolor{blue}{\underline{14.39}} & \textcolor{red}{\textbf{18.97}}
& 1264.45 & 1608.76
& \textcolor{blue}{\underline{16.65}} & 57.53
& 16.01 & 61.07 \\

DropPatch~\cite{droppatch}\!\!&\!\!\pub{AAAI25}
& 43.15 & 405.98
& 44.09 & 409.18
& 46.19 & 422.93
& 58.60 & 390.51
& 53.18 & 407.79
& 40.82 & 405.70 \\

WPMixer~\cite{murad2025wpmixer}\!\!&\!\!\pub{AAAI25}
& 17.20 & \textcolor{blue}{\underline{25.30}}
& 46.46 & 68.77
& 92.62 & 123.65
& 152.17 & 183.96
& 43.49 & 64.27
& 20.20 & 41.83 \\
\midrule

DASR~\cite{jeong2024decomposed}\!\!&\!\!\pub{KDD24}
& 15.67 & 29.59
& 31.73 & 49.66
& 127.21 & 159.01
& \textcolor{blue}{\underline{45.76}} & \textcolor{red}{\textbf{59.20}}
& 35.96 & 64.95
& 16.51 & 29.64 \\

KiGRU~\cite{Chu2025KiGRUFL}\!\!&\!\!\pub{SMC25}
& 17.60 & 32.00
& 37.05 & \textcolor{blue}{\underline{46.31}}
& 967.64 & 1209.55
& 66.92 & \textcolor{blue}{\underline{83.65}}
& 49.74 & 82.18
& 20.23 & 35.29 \\
\midrule

\multicolumn{2}{c}{\textsc{OrbitNet}~(Ours)}
& \textcolor{red}{\textbf{8.33}} & 27.46
& \textcolor{red}{\textbf{14.12}} & \textcolor{red}{\textbf{38.47}}
& \textcolor{red}{\textbf{14.37}} & \textcolor{blue}{\underline{19.75}}
& \textcolor{red}{\textbf{32.68}} & 106.97
& \textcolor{red}{\textbf{10.34}} & \textcolor{red}{\textbf{38.82}}
& \textcolor{red}{\textbf{4.55}} & \textcolor{red}{\textbf{10.00}} \\
\bottomrule
\end{tabular}
}
\end{table*}

We here evaluate the cross-constellation generalization ability of \textsc{OrbitNet} under the zero-shot setting, where the model is trained on the Starlink dataset and directly tested on unseen satellite constellations. The quantitative results are summarized in Table~\ref{tab:zero_shot_results}.

We observe that \textsc{OrbitNet} achieves the best MAE on all six datasets, demonstrating strong cross-constellation generalization ability.
In terms of RMSE, \textsc{OrbitNet} obtains the best results on ICEYE, LEMUR, and SKYSAT, and ranks second on CAPELLA, while DLinear~\cite{dlinear} and DASR~\cite{jeong2024decomposed} obtain lower RMSE on ASTROCAST and KINEIS, respectively.
More specifically, \textsc{OrbitNet} attains substantial gains on several challenging benchmarks.
On SKYSAT and ASTROCAST, it improves over the second-best baselines by $59.2\%$ and $32.5\%$ in MAE, respectively. Even when compared with large-scale time-series foundation models, \eg, TimesFM~\cite{das2023decoder} and MOIRAI~\cite{woo2024unified}, our model maintains a significant performance lead.
Compared with the domain-specific baselines, \textsc{OrbitNet} also shows stronger transferability across unseen constellations.
For example, DASR~\cite{jeong2024decomposed} and KiGRU~\cite{Chu2025KiGRUFL} obtain average MAE values of $45.47$ m and $193.20$ m across the six zero-shot datasets, respectively, while \textsc{OrbitNet} reduces the average MAE to $14.07$ m.

A further observation is that several generic forecasting baselines exhibit unstable behavior across constellations.
Although methods such as DLinear~\cite{dlinear} and iTransformer~\cite{itransformer} obtain competitive RMSE on certain datasets, their performance degrades sharply on others, especially on KINEIS, where the errors substantially increase.
The domain-specific baselines are generally more competitive than several generic methods, and DASR achieves the lowest RMSE on KINEIS.
However, they still show clear degradation on certain constellations, such as CAPELLA, where KiGRU reaches an MAE of $967.64$ m.
In contrast, \textsc{OrbitNet} produces more stable MAE performance across diverse satellite constellations. For instance, on SKYSAT, \textsc{OrbitNet} achieves an MAE of $4.55$ m, which is markedly lower than those of representative baselines such as AutoFormer, TimesNet, and KiGRU.
These results demonstrate the robustness and transferability of the proposed orbit-specific forecasting framework under cross-constellation zero-shot evaluation.

\subsection{Qualitative Results}
\label{subsec:qualitative_results}

\begin{figure*}[t]
\centering
\begin{subfigure}[t]{1.00\linewidth}
    \centering
    \includegraphics[width=\linewidth]{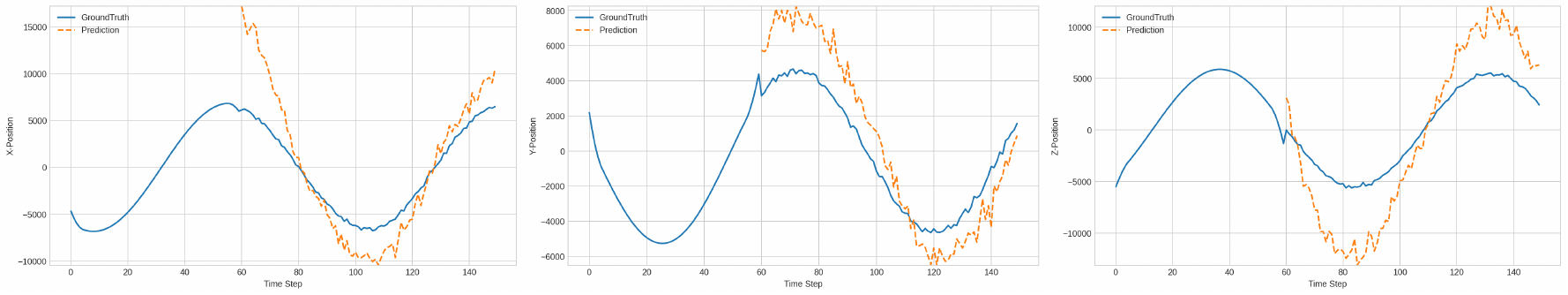}
    \caption{TimesNet~\cite{timesnet}}
\label{fig:qualitative_timesnet}
\end{subfigure}

\begin{subfigure}[t]{1.00\linewidth}
\centering
    \includegraphics[width=\linewidth]{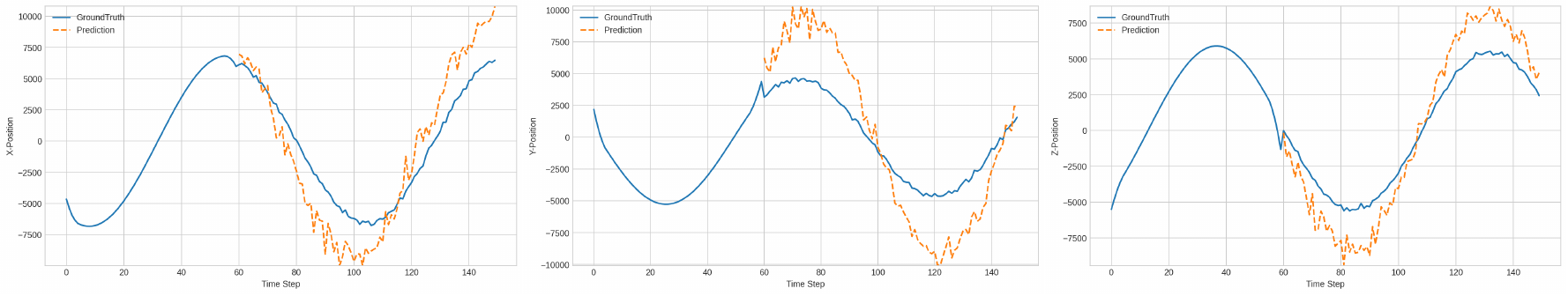}
    \caption{iTransformer~\cite{itransformer}}
\label{fig:qualitative_itransformer}
\end{subfigure}

\begin{subfigure}[t]{1.00\linewidth}
\centering
    \includegraphics[width=\linewidth]{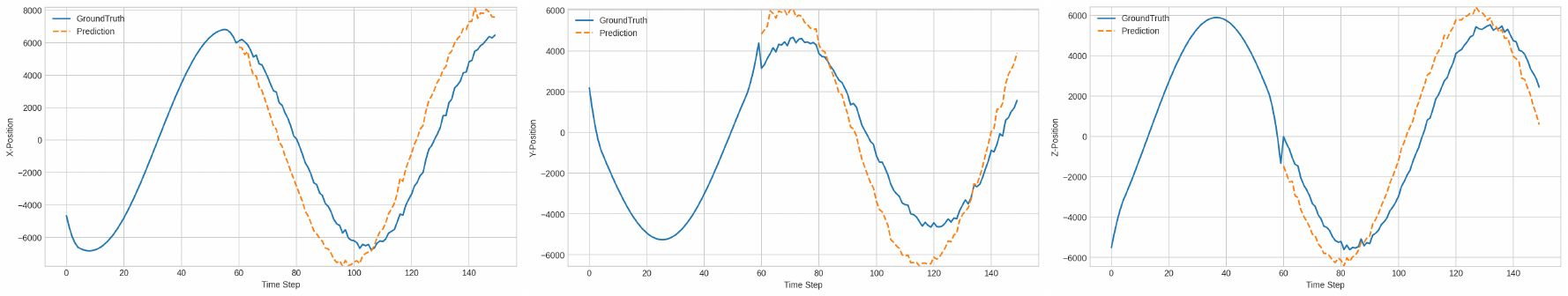}
    \caption{DLinear~\cite{dlinear}}
\label{fig:qualitative_dlinear}
\end{subfigure}

\begin{subfigure}[t]{1.00\linewidth}
\centering
    \includegraphics[width=\linewidth]{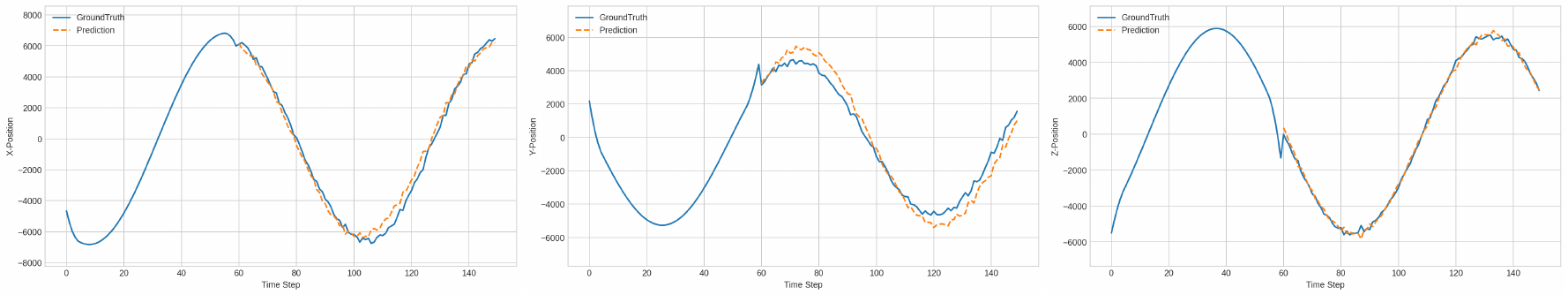}
    \caption{\textsc{OrbitNet} (Ours)}
\label{fig:qualitative_orbitnet}
\end{subfigure}

\caption{Qualitative results on Starlink dataset (Section~\ref{subsec:qualitative_results}). From top to bottom, we show the predicted results of TimesNet~\cite{timesnet}, iTransformer~\cite{itransformer}, DLinear~\cite{dlinear}, and \textsc{OrbitNet}. Each result is compared with the ground-truth trajectory along the three spatial coordinates.}
\Description{Four stacked trajectory prediction plots compare TimesNet, iTransformer, DLinear, and OrbitNet against ground truth on the Starlink dataset across the x, y, and z coordinates.}
\label{fig:qualitative_results}
\end{figure*}

The qualitative comparison is presented in Figure~\ref{fig:qualitative_results}, which visualizes the predicted orbital trajectories across three spatial coordinates for TimesNet~\cite{timesnet}, iTransformer~\cite{itransformer}, DLinear~\cite{dlinear}, and our \textsc{OrbitNet}.
It can be observed that \textsc{OrbitNet} produces predictions that are more closely aligned with the ground-truth trajectories than those of the competing methods.
Specifically, Figure~\ref{fig:qualitative_timesnet} shows that TimesNet~\cite{timesnet} exhibits noticeable deviations from the ground truth across all three coordinates.
Although it roughly captures the global trend, its predictions suffer from substantial offsets, especially around peaks and valleys.
A similar phenomenon can also be observed for iTransformer in Figure~\ref{fig:qualitative_itransformer}.
As shown in Figure~\ref{fig:qualitative_dlinear}, DLinear~\cite{dlinear} produces more stable predictions than the Transformer-based baselines and better preserves the overall periodic pattern, but clear discrepancies remain.
In contrast, Figure~\ref{fig:qualitative_orbitnet} shows that the predictions of \textsc{OrbitNet} almost overlap with the ground truth, demonstrating its ability to accurately capture both global trajectory trends and local orbital variations.

\subsection{Ablation Study}
\label{subsec:ablation_study}

In this section, we study the effectiveness of the core design of \textsc{OrbitNet} on Starlink dataset.
To this end, we construct a baseline model that takes only position sequence as input, denoted as ``\textit{Baseline}'' in Tables~\ref{tab:key_component_ablation} and~\ref{tab:fusion_strategy_ablation}.

\subsubsection{Key Component Analysis}
\label{subsubsec:key_component_analysis}

We first investigate the effectiveness of the two core components in \textsc{OrbitNet}, \ie, velocity-coupled representation refinement and orbital segment modeling.
The results are summarized in Table~\ref{tab:key_component_ablation}.
Three crucial observations can be drawn.
\textit{First}, velocity-coupled refinement brings noticeable improvements over the baseline on both MAE and RMSE. This indicates that incorporating velocity information through representation-level coupling provides more informative features for accurate orbit prediction.
\textit{Second}, we also observe compelling gains by incorporating the orbital segment modeling strategy into the baseline, decreasing MAE from $14.48$ m to $4.27$ m.
\textit{Third}, the full \textsc{OrbitNet}, which integrates both components, achieves the best performance, with an MAE of $3.34$ m and an RMSE of $22.80$ m, which confirms that the two components are complementary.

\subsubsection{Velocity-coupled Representation Refinement}
\label{subsubsec:fusion_strategy_analysis}

\begin{table}[t]
\centering

\begin{minipage}[t]{0.48\textwidth}
\centering
\caption{Ablation study of the core components of \textsc{OrbitNet} on the Starlink dataset (Section~\ref{subsubsec:key_component_analysis}).}
\label{tab:key_component_ablation}
\tabstyle{0.85}{4}{1.2}{
\begin{tabular}{lcc}
    \toprule
    Variant & MAE & RMSE \\
    \midrule
    \textit{Baseline} & 14.48 & 31.94 \\
    \quad + Velocity Refinement & 5.43 & 29.83 \\
    \quad + Segment Modeling & 4.27 & 27.39 \\
    \quad + Full (\textsc{OrbitNet}) & 3.34 & 22.80 \\
    \bottomrule
\end{tabular}
}
\end{minipage}
\hfill
\begin{minipage}[t]{0.48\textwidth}
\centering
\caption{Ablation study of velocity fusion strategies in velocity-coupled refinement (Section~\ref{subsubsec:fusion_strategy_analysis}).}
\label{tab:fusion_strategy_ablation}
\tabstyle{0.90}{4}{1.2}{
\begin{tabular}{lccc}
    \toprule
    Input & Strategy & MAE & RMSE \\
    \midrule
    \textit{Baseline} & -- & 14.48 & 31.94 \\ 
    Corr~\textit{Only} & -- & 91.56 & 136.93 \\
    \midrule
    \multirow{3}{*}{Position + Velocity} 
    & \texttt{Concat} & 10.53 & 51.19 \\
    & \texttt{Sum} & 8.73 & 34.06 \\
    & \texttt{Coupling} & 3.34 & 22.80 \\
    \bottomrule
\end{tabular}
}
\end{minipage}

\end{table}

We next examine the design of velocity-coupled refinement. In Table~\ref{tab:fusion_strategy_ablation}, ``Corr~\textit{Only}'' refers to the variant that uses only the correlation-aware representation $\mathbf{H}_{\mathrm{corr}}$ for prediction, without retaining the original position sequence.
As shown, the ``Corr~\textit{Only}'' variant performs worse than the \textit{Baseline}, with MAE/RMSE increasing from $14.48/31.94$ m to $91.56/136.93$ m.
This result indicates that the discriminative information for orbit prediction still comes from the position sequence itself, while velocity-related information alone cannot provide a reliable basis for precise forecasting.
Furthermore, when both position and velocity are used, the choice of fusion strategy has a pronounced impact on performance. 
Specifically, the simple \texttt{concat} strategy yields the weakest results, while \texttt{sum} shows clear improvement, achieving an MAE of $8.73$ m and an RMSE of $34.06$ m.
By contrast, the proposed \texttt{coupling} strategy achieves the best performance and consistently outperforms all other settings.

\subsubsection{Coefficient $\alpha$ Analysis}
\label{subsubsec:fusion_coefficient_analysis}

Table~\ref{tab:fusion_coefficient_ablation} quantifies the effect of the fusion coefficient $\alpha$ in Eq.~\eqref{eq:couple}, which controls the relative contribution of the original position sequence and the velocity-coupled representation.
As shown, the prediction performance varies substantially under different fixed values of $\alpha$.

As shown in Table~\ref{tab:fusion_coefficient_ablation}, the prediction performance progressively improves as $\alpha$ is increased from $0$ to $1$.
In particular, when $\alpha{=}0$, the model relies solely on the velocity-coupled representation and leads to unsatisfactory results.
As $\alpha$ increases, more original positional information is retained, and the MAE decreases to $4.29$ m when $\alpha=1$.
Notably, the learnable $\alpha$ achieves the best performance, further reducing the MAE and RMSE to $3.34$ m and $22.80$ m, respectively.
These results demonstrate that a fixed fusion coefficient is suboptimal, whereas adaptively balancing the original positional signal and the velocity-coupled representation enables the model to exploit more informative orbital features.

\subsubsection{Position Embedding Analysis}
\label{subsubsec:position_embedding_analysis}

\begin{table}[t]
\centering

\begin{minipage}[t]{0.48\textwidth}
\centering
\caption{Ablation study of the fusion coefficient $\alpha$ in velocity-coupled refinement (Section~\ref{subsubsec:fusion_coefficient_analysis}).}
\label{tab:fusion_coefficient_ablation}
\tabstyle{0.85}{6}{1.2}{
\begin{tabular}{ccc}
\toprule
    Coefficient $\alpha$ (Eq.~\ref{eq:couple}) & MAE & RMSE \\
    \midrule
    $\alpha=$0.00 & 91.56 & 136.93 \\
    $\alpha=$0.25 & 8.42 & 40.50 \\
    $\alpha=$0.50 & 4.89 & 27.02 \\
    $\alpha=$1.00 & 4.29 & 26.27 \\
    Learnable & 3.34 & 22.80 \\
    \bottomrule
\end{tabular}
}
\end{minipage}
\hfill
\begin{minipage}[t]{0.48\textwidth}
\centering
\caption{Ablation study of positional embedding designs in orbital segment modeling (Section~\ref{subsubsec:position_embedding_analysis}).}
\label{tab:position_embedding_ablation}
\tabstyle{0.95}{8}{1.2}{
\begin{tabular}{cccc}
\toprule
    \multirow{2}{*}{Metric} & \multirow{2}{*}{\textit{w/o} PE} & \multicolumn{2}{c}{\textit{w}/ PE} \\
    \cmidrule(lr){3-4}
    & & Sinusoidal & Learnable \\
    \midrule
    MAE & 6.95 & 5.29 & 3.34 \\
    RMSE & 30.50 & 26.65 & 22.80 \\
    \bottomrule
\end{tabular}
}
\end{minipage}

\end{table}

Furthermore, we investigate the impact of positional embedding (PE) designs within the orbital segment modeling.
As shown in Table~\ref{tab:position_embedding_ablation}, introducing positional embedding leads to substantial performance gains over the variant without positional embedding (\ie, \textit{w/o} PE).
To further confirm that the improvements come from encoding segment-level temporal order, we consider an alternative approach using a fixed sinusoidal positional embedding. 
The results show that the sinusoidal design also boosts performance over \textit{w/o} PE, reducing MAE from $6.95$ m to $5.29$ m.
Moreover, learnable positional embedding achieves better performance, suggesting that adaptive temporal order encoding is more suitable for orbital segment modeling.

\subsubsection{Hyperparameter Analysis}
\label{subsubsec:hyperparameter_analysis}

\begin{figure}[t]
\centering

\begin{subfigure}[t]{0.48\linewidth}
    \centering
    \includegraphics[width=\linewidth]{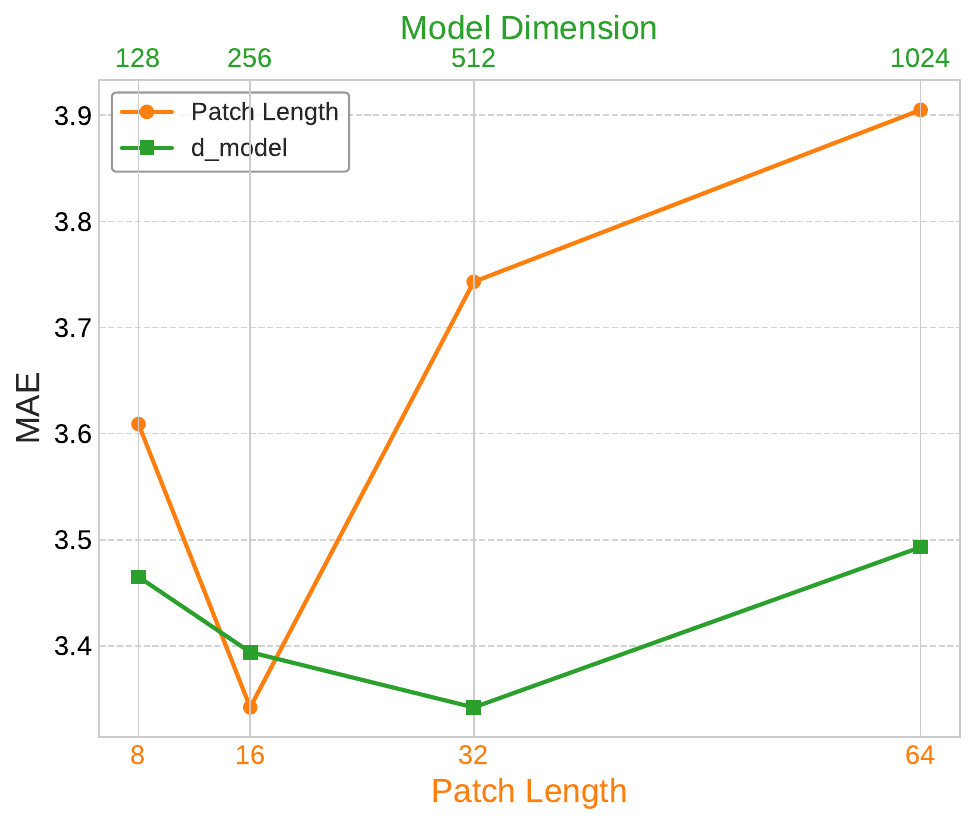}
    \caption{Hyperparameter analysis}
    \label{fig:hyperparameter_analysis}
\end{subfigure}
\hfill
\begin{subfigure}[t]{0.48\linewidth}
    \centering
    \includegraphics[width=\linewidth]{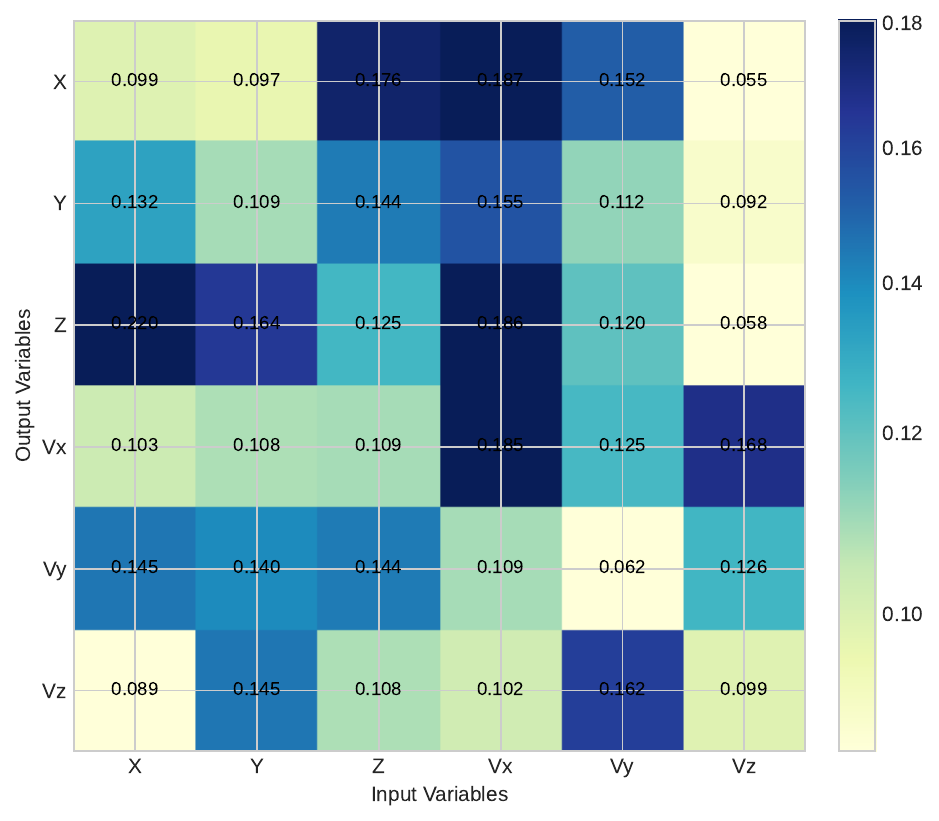}
    \caption{Variable dependency matrix}
    \label{fig:variable_correlation_analysis}
\end{subfigure}

\caption{Analysis of \textsc{OrbitNet}. (a) Hyperparameter analysis of patch length $L_p$ in Eq.~\eqref{eq:patch} and hidden dimension $d$ in Eq.~\eqref{eq:project}. (b) Variate-wise dependency matrix learned in velocity-coupled refinement.}
\Description{Two-panel analysis figure showing line plots of mean absolute error under different patch lengths and hidden dimensions, and a heatmap of the learned dependency matrix among position and velocity variables.}
\label{fig:analysis_results}
\end{figure}

Figure~\ref{fig:hyperparameter_analysis} illustrates the effect of two key hyperparameters in~\textsc{OrbitNet}, \ie, the patch length $L_p$ in Eq.~\eqref{eq:patch} and the hidden dimension $d$ in Eq.~\eqref{eq:project}.
We observe that the patch length has a noticeable impact on prediction accuracy.
The MAE decreases substantially when $L_p$ increases from $8$ to $16$, but rises again when the patch length is further enlarged to $32$ and $64$.
In addition, the MAE decreases as the hidden dimension $d$ increases from $128$ to $512$, reaching the best performance at $d=512$, but degrades when $d$ is further increased to $1024$.
Hence, we set $L_p{=}16$ and $d{=}512$ as the default hyperparameter configuration.

\subsection{Application to Long-term Orbit Prediction}
\label{subsec:long_term_prediction}

\begin{table}[t]
\centering

\caption{Long-term forecasting results under different prediction horizons (Section~\ref{subsec:long_term_prediction}).}

\label{tab:long_term_results}
\tabstyle{0.52}{8}{1.2}{
\begin{tabular}{lccc}
\toprule
Method & Horizon & MAE & RMSE \\
\midrule

\multirow{3}{*}{TimesFM~\cite{das2023decoder}}
& 180 & 48.57 & 61.81 \\
& 360 & 71.58 & 94.92 \\
& 720 & 95.15 & 131.45 \\
\midrule

\multirow{3}{*}{iTransformer~\cite{itransformer}}
& 180 & \textcolor{blue}{\underline{10.91}} & \textcolor{blue}{\underline{31.30}} \\
& 360 & \textcolor{blue}{\underline{15.93}} & \textcolor{blue}{\underline{40.50}} \\
& 720 & \textcolor{blue}{\underline{17.77}} & \textcolor{blue}{\underline{36.88}} \\
\midrule

\multirow{3}{*}{DropPatch~\cite{droppatch}}
& 180 & 36.01 & 52.15 \\
& 360 & 205.48 & 272.80 \\
& 720 & 78.20 & 153.95 \\
\midrule

\multirow{3}{*}{\textsc{OrbitNet}~(Ours)}
& 180 & \textcolor{red}{\textbf{3.78}} & \textcolor{red}{\textbf{25.30}} \\
& 360 & \textcolor{red}{\textbf{6.30}} & \textcolor{red}{\textbf{27.93}} \\
& 720 & \textcolor{red}{\textbf{10.81}} & \textcolor{red}{\textbf{31.94}} \\
\bottomrule
\end{tabular}
}
\end{table}

Long-term prediction is essential in satellite orbit analysis, especially for downstream applications such as collision avoidance and space traffic management.
To further evaluate the applicability of \textsc{OrbitNet} beyond the default forecasting horizon, we extend the prediction length to $180$, $360$, and $720$ steps and compare it with selected representative baselines.
The experimental results are listed in Table~\ref{tab:long_term_results}.
As illustrated, longer forecasting horizons generally lead to larger prediction errors, which can be attributed to error accumulation over extended future steps.
Nevertheless, our \textsc{OrbitNet} achieves the best performance across all three long-term settings among the selected baselines.
These numerical results indicate that \textsc{OrbitNet} is able to preserve relatively stable prediction quality under longer forecasting horizons, showcasing its effectiveness for long-term orbit prediction.

\subsection{Variable Correlation Analysis}
\label{subsec:correlation_analysis}

To further examine how \textsc{OrbitNet} captures cross-variable interactions, we analyze the convolution operator in the proposed velocity-coupled refinement.
Since the operator applies one-dimensional convolution to the concatenated orbital state sequence $[\mathbf{I}_{\mathrm{pos}},\mathbf{I}_{\mathrm{vel}}]$, its learned weights can provide clues about dependencies among the six input variables, \ie, $(x,y,z,v_x,v_y,v_z)$.
Accordingly, we aggregate the absolute convolution weights along the kernel dimension to obtain a variate-wise dependency matrix, as shown in Figure~\ref{fig:variable_correlation_analysis}.

Two main observations can be drawn from Figure~\ref{fig:variable_correlation_analysis}.
\textit{First}, the learned matrix exhibits clear non-diagonal responses, indicating that the refinement module does not process the orbital variables independently, but instead learns cross-variable interactions.
\textit{Second}, it reveals notable dependencies between position and velocity variables, such as $x \leftarrow v_x$ ($0.187$), $z \leftarrow v_x$ ($0.186$), and $x \leftarrow v_y$ ($0.152$), suggesting that velocity variables provide useful cues for refining positional representations.
At the same time, the matrix also exhibits dependencies within position variables and within velocity variables, such as $z \leftarrow x$ ($0.220$) and $v_z \leftarrow v_y$ ($0.162$).
Hence, these findings provide interpretability support for the proposed velocity-coupled representation refinement and suggest that it learns meaningful dependencies across orbital variables for informative representation learning.

\section{Conclusion and Discussion}
\label{sec:conclusion}

\subsection{Conclusion}
\label{subsec:conclusion}

In this paper, we study satellite orbit prediction from a sequence learning perspective and identify an important limitation of existing learning-based approaches, \ie, insufficient exploitation of the intrinsic relationship between position and velocity.
To address this limitation, we propose \textsc{OrbitNet}, a task-oriented forecasting framework for accurate satellite orbit prediction.
At its core, \textsc{OrbitNet} integrates velocity-coupled refinement and orbital segment modeling. The former enhances positional representations by injecting velocity-aware motion cues, while the latter captures orbital evolution patterns through segment-level temporal representation learning.
Extensive experiments demonstrate that \textsc{OrbitNet} achieves the best in-domain performance and the best zero-shot MAE across six unseen satellite constellations, while maintaining competitive RMSE on most zero-shot datasets.
These results suggest that explicitly exploiting the structured relationship between position, velocity, and temporal orbital segments is critical for learning-based orbit prediction.

\subsection{Discussion}
\label{subsec:discussion}

Despite the encouraging results, two points merit further discussion:

\begin{itemize}

    \item \textit{Limited physical priors.}
    The current framework is mainly built on data-driven sequence learning and does not explicitly incorporate richer orbital mechanics priors, environmental perturbation factors, or control-related information such as maneuver events. Although \textsc{OrbitNet} learns temporal patterns from historical state sequences, its integration with explicit physical mechanisms remains limited. A tighter combination of sequence modeling and orbital dynamics priors may further improve physical consistency and long-horizon stability.

    \item \textit{Toward orbit foundation models.}
    While this work focuses on satellite orbit prediction, it also suggests the potential of developing foundation models for orbital time series. Satellite state sequences exhibit continuous temporal structures, shared dynamical regularities across constellations, and intrinsic position-velocity coupling, making them promising candidates for large-scale representation learning. In the future, a more general orbit foundation model may support not only trajectory prediction, but also broader downstream tasks such as uncertainty estimation, anomaly detection, conjunction risk assessment, and space traffic management.

\end{itemize}












\bibliographystyle{ACM-Reference-Format}
\bibliography{sample-base}


\end{document}